\pdfoutput=1
\documentclass[runningheads]{llncs}
\usepackage{graphicx}
\usepackage{comment}
\usepackage{amsmath,amssymb} % define this before the line numbering.
\usepackage{color}
\usepackage{cite}

\begin{document}
% \linenumbers
\pagestyle{headings}
\mainmatter
\def\ECCVSubNumber{5535}  % Insert your submission number here

\title{Redesigning Fully Convolutional DenseUNets \\ for Large Histopathology Images} % Replace with your title

\begin{comment}
\titlerunning{ECCV-20 submission ID \ECCVSubNumber}  % 
\authorrunning{ECCV-20 submission ID \ECCVSubNumber}  %
\author{Anonymous ECCV submission}
\institute{Paper ID \ECCVSubNumber}
\end{comment}
%******************

%\begin{comment}
\titlerunning{Redesigning Fully Convolutional DenseUNets}
\author{Juan P. Vigueras-Guill\'en\inst{1} \and % \orcidID{0000-0002-5676-8217}
Joan Lasenby\inst{2} \and %\orcidID{0000-0002-0571-0218}
Frank Seeliger\inst{1}} % \orcidID{0000-0003-0565-951X}
\authorrunning{Vigueras-Guill\'en et al.}
\institute{CVRM Safety, Clinical Pharmacology and Safety Science, R\&D, AstraZeneca, Gothenburg, Sweden, \and
Department of Engineering, University of Cambridge, Cambridge, UK.\\
\email{JuanPedro.ViguerasGuillen@astrazeneca.com}}
%\end{comment}
%******************

\maketitle

\begin{abstract}
The automated segmentation of cancer tissue in histopathology images can help clinicians to detect, diagnose, and analyze such disease. Different from other natural images used in many convolutional networks for benchmark, histopathology images can be extremely large, and the cancerous patterns can reach beyond 1000 pixels. Therefore, the well-known networks in the literature were never conceived to handle these peculiarities. In this work, we propose a Fully Convolutional DenseUNet that is particularly designed to solve histopathology problems. We evaluated our network in two public pathology datasets published as challenges in the recent MICCAI 2019: binary segmentation in colon cancer images (DigestPath2019), and multi-class segmentation in prostate cancer images (Gleason2019), achieving similar and better results than the winners of the challenges, respectively. Furthermore, we discussed some good practices in the training setup to yield the best performance and the main challenges in these histopathology datasets. \footnote{This work was originally submitted to ECCV 2020.}

\end{abstract}

\section{Introduction}

Fully Convolutional Networks (FCNs) \cite{Long15, Ronneberger15} were introduced as a way to expand Convolutional Neural Networks (CNNs) for semantic image segmentation, where the neural network (NN) layers at the end of the CNN were substituted with an upsampling path to recover the spatial resolution of the input image. One major contribution in these networks was the introduction of skip-connections between downsampling and upsampling paths \cite{Ronneberger15}, which have proven to be effective in recovering fine-grained details of the images \cite{Drozdzal16}. 

Many CNNs architectures have been extended to FCNs. This is the example of the so-called Tiramisu network \cite{Jegou17}, which uses the design of DenseNets \cite{Huang17}. DenseNets exploit the idea of dense connections, which consists of the concatenation of all previous feature maps (within a resolution/dense block) before sending them to the next convolutional layer. This concept could be interpreted as the use of skip-connections within a dense block, and it shows many similarities with another well-known architecture named Residual Networks (ResNets) \cite{He2016}, although ResNets sum the feature maps instead of concatenate them. Overall, DenseNets have shown many benefits: (1) features are reused along a resolution block, where all convolutional layers can see the preceding feature maps; (2) there is an implicit deep supervision; (3) the network is robust against overfitting \cite{ViguerasSPIE19}; (4) since the information of the gradients or input is sent through the short-connections, it avoids the vanishing gradient problem \cite{He2016}.

When aiming for semantic segmentation in histopathology problems, images are usually much larger than the ones in the public datasets used for benchmark, like ImageNet, whose images are 256$\times$256 pixels. Since current GPUs do not have enough memory to handle large images, it is common to subdivide the image in patches and solve them independently. However, this raises the questions of how large the patches should be and what would be the correct balance between network size (in parameters) and patch size. We hypothesize that, for best results, a patch should cover the whole area where abnormal, growing cancer tissue occurs such that the network could understand the limits of the cancerous tissue.

In this work, we propose a Fully Convolutional DenseNet designed for histopathology images. While many details in the Tiramisu network \cite{Jegou17} can be optimal for small images, they might require many computational resources and their contribution might not be relevant. Thus, we redesigned FC-DenseNets for better performance in large pathology images. To prove the potential of our network, we experimented with two public histopathology datasets presented as challenges in the International Conference on Medical Image Computing and Computer Assisted Intervention (MICCAI 2019): (1) a dataset of colon cancer, named DigestPath2019, to perform binary segmentation; and (2) a dataset of prostate cancer, named Gleason2019, to perform grading (multi-class segmentation).

The contributions of this work are as follows: (1) we propose a new Fully Convolutional DenseNet designed for large histopathology images; (2) we evaluate the method in two different datasets and compare them with the well-known Tiramisu network \cite{Jegou17}; (3) we discuss the training details that are key for a good performance, such as class balance and image sampling.

\section{Methods}

\subsection{Datasets}

\subsubsection{DigestPath2019.}
\begin{figure}[t]
	\centering
	\includegraphics[width=1\linewidth]{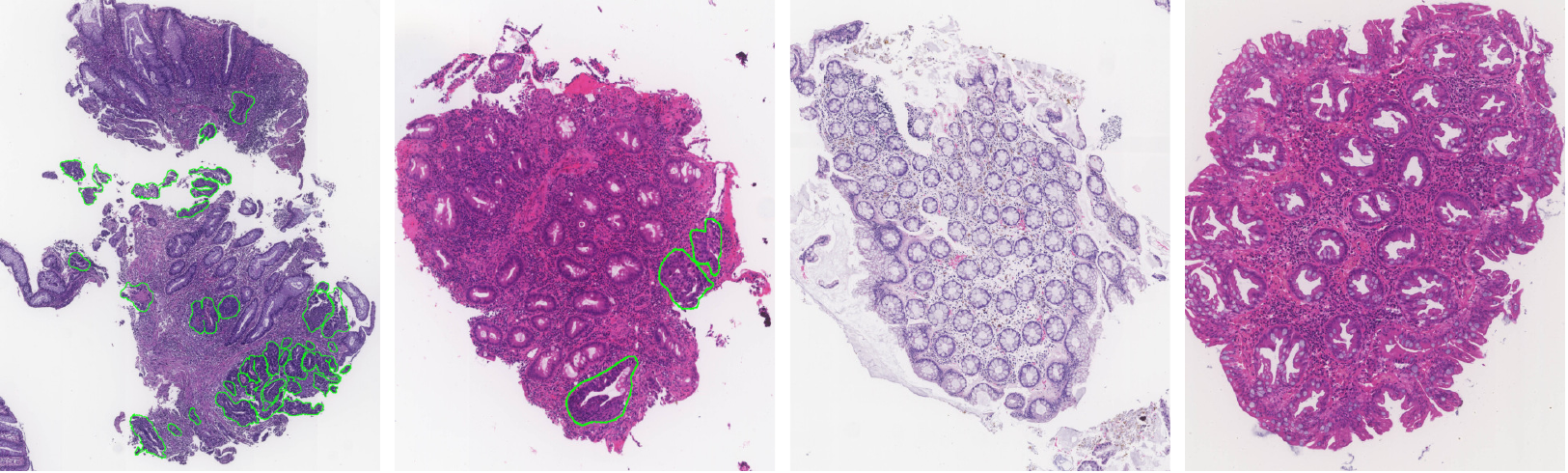}
	\caption{Four representative examples of the DigestPath dataset; from left to right, two positive images (the perimeter of the lesions are highlighted in green), and two negatives images. These images were scaled for illustrative purposes (leftmost image had a size 5953$\times$7294 pixels whereas rightmost image had a size of 3312$\times$4096).}
	\label{fig01}
\end{figure}

The `colonoscopy tissue segment dataset' was part of the DigestPath2019 Challenge \cite{DigestPath19}. According to the authors, 660 color tissue slices of an average size of 5000x5000 pixels from 324 patients were provided as training, from which 250 images had a lesion annotation (positive images) and the remaining 410 contained healthy tissue (negative images). This was a very unbalanced two-class problem: the pixel ratio between the malignant tissue (positive class) and healthy tissue or non-tissue (negative class) within the positive images was 1:8, which increased to 1:14 if the negative images were considered. There was a single annotation per image, although it was not specified whether the same pathologist annotated all images. The data showed large variations in terms of appearance because it was collected from several medical centers in developing countries (Fig.~\ref{fig01}). All whole slide images were stained with hematoxylin and eosin (H\&E) and scanned at X20. The malignant lesions in the dataset were high grade intraepithelial neoplasia and adenocarcinoma, including papillary adenocarcinoma, mucinous adenocarcinoma, poorly cohesive carcinoma and signet ring cell carcinoma. 

The testing dataset was not released. Thus, we performed a 10-fold cross-validation in the training set, where fold 1 was considered the validation set used to compare the different networks. Once we established the best network, the remaining 9 folds were cross-validated and the final test values were computed as the average between the 10 folds. To evaluate the segmentation results, the metrics chosen was accuracy and Dice Similarity Coefficient (DICE), where the latter measures the area overlap between segmentation results and annotations and is defined as
\begin{align}
	\text{DICE} =  \frac{2 \left| A \cap B \right|}{\left| A \right| + \left| B \right|} \times 100\%,
\end{align}
where $A$ is the set of foreground pixels in the annotation and $B$ is the set of foreground pixels in the segmentation result. More than 500 participants registered to the challenge, and the winner obtained a $\text{DICE}=80.75\%$. 

Overall, the quality of the annotations was good, with a precise delineation of the tumors. However, we noted that some non-tissue areas within tumors were annotated as cancer pixels (Fig.~\ref{fig01}). While this is an understandable decision if the annotations were used for discussion between pathologists, it seems rather counterproductive for an algorithm and it raises the question of whether a CNN could understand those human-made patterns in the labels. In summary, the two main challenges in this dataset were the high class imbalance and the diverse appearance of the images (color, contrast, etc.). 

\subsubsection{Gleason2019.}
\begin{figure}[t]
	\centering
	\includegraphics[width=1\linewidth]{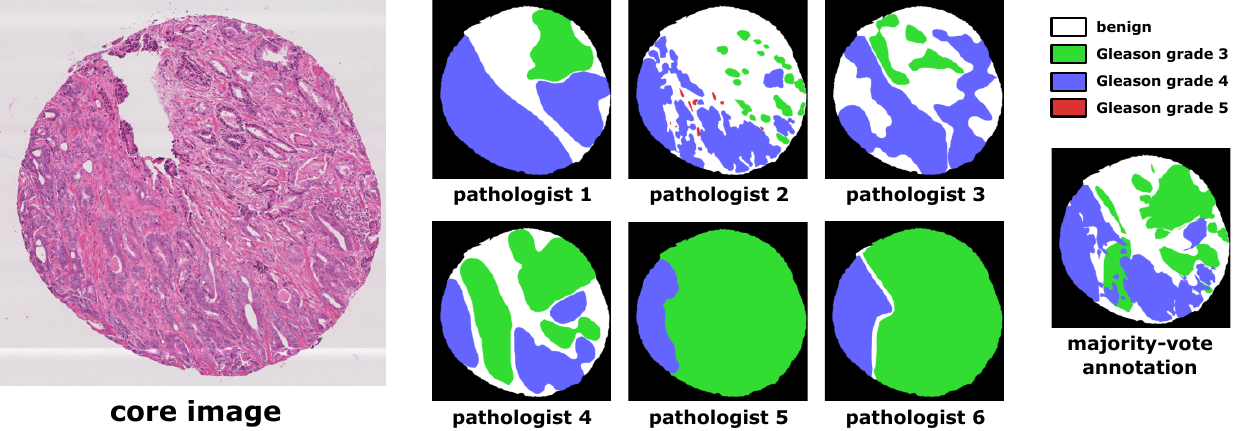}
	\caption{One representative example of the pathologists' annotations in the Gleason Challenge. The majority-vote label is the one used to evaluate the models.}
	\label{fig02}
\end{figure}

This challenge dealt with the automatic Gleason grading of prostate cancer from H\&E-stained histopathology images \cite{Gleason19}. The Gleason grading is a 5-grade system named after the pathologist who developed it in the 1960s, Dr. Donald Gleason, and indicates the distinct patterns as the cancerous cells change from normal to tumor cells \cite{GleasonScore19}, where grade 1 refers to cancer cells that resemble normal prostate tissue and grade 5 indicates highly mutated cancer cells. This dataset contained 244 training images of an average size of 5120$\times$5120 pixels, and six pathologists performed the grading annotations, although not in all images. Exactly, the six pathologists performed 242, 136, 238, 240, 244, and 65 annotations, respectively (a total of six annotations were discarded due to clear mistakes). In this challenge, the goal was to segment the grades 3, 4, and 5; thus, labels 1 and 2 were merged with label 0 (healthy tissue and non-tissue). To obtain the gold standard labels, the authors defined its computation as the pixel-based majority voting among the available annotations, although they did not disclose these majority-vote labels and did not explain how to proceed when a tie occurs or whether labels 1 and 2 were considered for the majority vote and later discarded or vice versa. Therefore, we decided to first remove labels 1--2 and then compute the majority-vote labels, setting the higher grade in case of a tie (Fig.~\ref{fig02}). 

When observing the annotations, we noted that most of the annotators made a rough, imprecise work (Fig.~\ref{fig02}). Indeed, each one of the pathologists showed a different way of annotating: pathologist \#2 was the exception by being very detailed oriented; pathologists \#5 and \#6 tended to label the whole tissue area (\#5 barely left any area unlabeled, whereas the others liked to leave unlabeled stripes between labeled areas), and pathologists \#1, \#3 and \#4 were somewhere in between. Furthermore, the labeled areas sometimes reached beyond the tissue, which suggests that the pathologists did not pay attention to the tissue borders and preferred to do a quick job. To mitigate this problem, we automatically created binary masks differentiating the tissue area from the outer-area and we applied it to the label images, thus removing any label wrongly applied outside the tissue area. This also helped to, later in the training, extract patches from the healthy tissue (white pixels in Fig.~\ref{fig02}) and not from the outer-area (black pixels in Fig.~\ref{fig02}).

The labels of the test set were not released. Thus, we also created a 10-fold cross-validation in the training set and performed exactly as described in the DigestPath dataset. To evaluate the segmentation results, the metric chosen by the authors was a combination of Cohen's kappa and the F1-scores. Specifically, the metric was defined as

\begin{align}
	\text{score} =  \text{Cohen's kappa} + \frac{\text{macro-averaged F1} + \text{micro-averaged F1}}{2},
\end{align}
where Cohen's kappa expresses the level of agreement between two annotators, computed as $\kappa = (p_o - p_e)/(1-p_e)$, being $p_o$ the observed agreement ratio and $p_e$ the expected agreement when both annotators assign labels randomly. F1-score is computed as $\text{F1} = 2 \cdot (precision \cdot recall)/(precision + recall)$, being $precision = \text{TP} / (\text{TP} + \text{FP})$ and $recall = \text{TP} / (\text{TP} + \text{FN})$, using the global computation of true positives (TP), false negatives (FN) and false positives (FP) to yield the `micro F1', and using the computation for each label to obtain the `macro F1'. We used the implementation of this formulas from the Scikit-learn python API \cite{scikitLearn}. Finally, more than 100 participants registered to this challenge, and the winner obtained a $\text{score}=0.8451$. 

Overall, this was a more complex problem than the DigestPath challenge. For instance, the cases where non-tissue inner-areas were labeled as cancer occurred more often in this dataset, with extreme cases such as the one displayed in the upper part of the core image in Fig.~\ref{fig02}. The grades were also highly unbalanced, being the proportion of grades in the majority-vote labels $[25, 22, 36, 1]$ for healthy-tissue (including grades 1 and 2) and grades 3, 4, and 5, respectively; this is, for one pixel with grade 5, there were 36 pixels with grade 4. In summary, the main challenges in this dataset were the high class imbalance, the poorly annotated labels, and the discrepancies of opinions between pathologists.

\subsection{Network}

Our proposed network is depicted in Fig.~\ref{fig03}, named 1BN-DenseUnet, and it could be seen as a different adaptation of DenseNets \cite{Huang17} for semantic segmentation than the Tiramisu network \cite{Jegou17}. Many changes were particularly suitable for our histopathology problems and, thus, they might not be extrapolated to other type of natural images. These were:

\begin{figure}[t]
	\centering
	\includegraphics[width=1\linewidth]{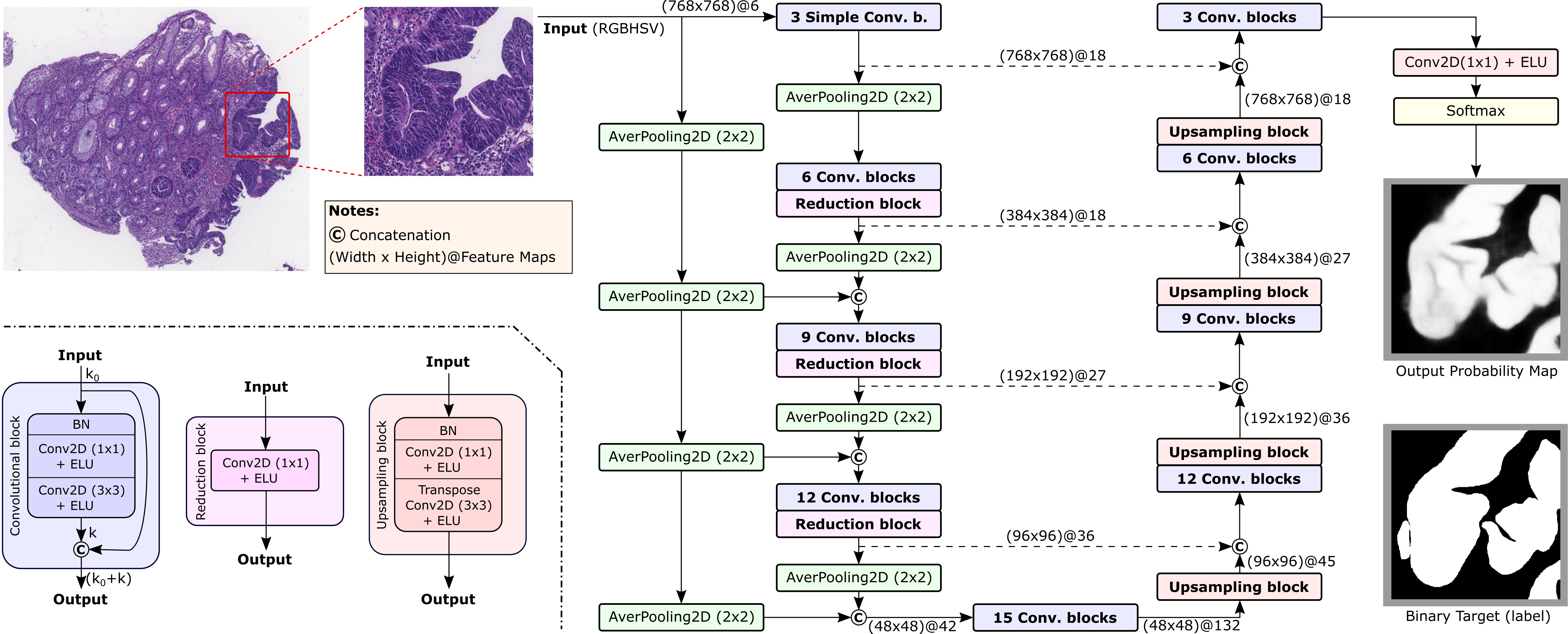}
	\caption{Schematic overview of the 1BN-DenseUnet network.  }
	\label{fig03}
\end{figure}

\begin{enumerate}
	\item The main hyperparameters were set so that the network would fit within our limited resources (GPU with 16GB RAM). This involved a growth rate of GR=6 (feature maps at each convolutional layer Conv2D(3$\times$3)) and dense blocks with the following number of convolutional blocks $[3,6,9,12,15]$.
	\item Different from both \cite{Huang17, Jegou17}, we employed the ELU function activation \cite{Clevert17} instead of ReLUs \cite{Glorot11}, and they were set within the convolutional layer simply to save resources.
	\item Batch Normalization (BN) \cite{Ioffe2015} layers were only used before the feature reduction layer, Conv2D(1$\times$1). As later discussed, BN layers occupied many memory resources and could be reduced in number without decrease in performance.
	\item Similar to \cite{Huang17}, we used feature reduction layers, BN+Conv2D(1$\times$1)+ELU, with the same feature outputs of 4$\times$GR as in \cite{Huang17}. We used these layers in both, convolutional and upsampling blocks. In contrast, Tiramisu network does not use them.	
	\item Instead of a initial convolutional layer with larger filters as in \cite{Huang17, Jegou17}, we set a first dense block with simply three convolutional layers (no BN or feature reduction layer).
	\item There is no reduction block in the first dense block. We observed that having the input image in the last dense block helped with the fine-grained details.
	\item Different from \cite{Jegou17}, we up-sampled with all the concatenated features from the output of the previous dense block. Since we applied feature reduction before upsampling, the memory demanding was similar to Tiramisu, which only uses the output of the last convolutional block.
	\item Different from both \cite{Huang17, Jegou17}, we used a compression rate of C=0.5 in the reduction and upsampling block but only considering the number of features created in the previous dense block, computed as $\text{C} \times \text{GR} \times (\text{No. conv. blocks})$. In practice, we applied higher compression than \cite{Huang17, Jegou17}, reducing the feature maps to approximately 1/3 in the deepest layer.	
	\item New to both \cite{Huang17, Jegou17}, our network added the input image at the beginning of each dense block in the downsampling path (properly reduced in size). This idea has been previously suggested \cite{Zeng2018} and here it boosted the performance with just a very small increase in computational cost. 
	\item New to both \cite{Huang17, Jegou17}, our network introduced the images with two color spaces, RGB+HSV. The specific use of a color space is mainly data and problem related. In our case, we observed that adding both spaces helped the network to yield a slightly better accuracy than any of the two spaces independently and, most interestingly, the training showed a smoother loss and accuracy curve, suggesting a better convergence. 
\end{enumerate}

For comparison purposes, we implemented the Tiramisu network \cite{Jegou17}, setting the same depth and width as in Fig.~\ref{fig03}, yielding 329K parameters. In contrast, our network had 294K parameters.

These networks were employed in both datasets: in the DigestPath2019 we used a 2-class output (binary class), whereas in the Gleason2019 we employed a one-hot encoding and a probability encoding.

\subsection{Training Details}

\subsubsection{DigestPath2019.}
This dataset had two types of images (positives and negatives) and, within the positive images, two classes. To balance the classes, we extracted four patches of 756$\times$756 pixels to build the batch in the following way: two patches from the positive images whose center pixel was class 1 (cancer), one patch from the positive images whose center pixel was class 0 (non-cancer), and one patch from the negative images. Importantly, each patch was obtained from a different image, the non-cancer patches were extracted from tissue areas, and the patches were shuffled within the batch to avoid any learning bias. While this setting did not ensure a perfect 50/50 class balance in the batches, we observed that any attempt to further weight the classes did not improve the performance. Furthermore, the fact that we only sampled one patch from the negative images did not diminish the performance in that subset. 

For data augmentation, we performed only flipping and four rotations over angles of $k \cdot 90^{\circ}, k \in \{0,1,2,3\}$ in the batch. Thus, eight orientations were possible without introducing interpolation. The remaining hyperparameters were: binary cross-entropy as loss function, nadam optimizer \cite{Dozat2016}, 400 iterations for one epoch (approximately the number of negative images), 250 epochs, a initial learning rate of $lr=0.001$, and a rate decay of $lr_{decay}=0.99$, being the updated rate at each new epoch $lr_{new}= lr\times (lr_{decay})^{epoch}$. Since overfitting was not observed, we employed the last model (no early-stopping).

\subsubsection{Gleason2019.}
In this problem, we had four classes, with one heavily unrepresented (grade 5). Thus, we balanced the classes by constructing a batch with an example of each class. Similar to DigestPath2019, each patch was obtained from a different image, we extracted the patches for the non-cancer class from the healthy tissue (white pixels in Fig.~\ref{fig02}), and we also shuffled the patches within the batch. The key aspect in this problem was deciding how to combine the different pathologists' annotations, knowing that the majority-vote would be used for evaluation. Thus, we performed several experiments:

\begin{itemize}
	\item We trained using only the annotations by one pathologist (all of them were tested), using one-hot encoding.
	\item We trained with the majority-vote (one-hot encoding).
	\item We trained with all available annotations; thus we would randomly select a pathologist's annotation for each patch (one-hot encoding).
	\item We trained with a probabilistic encoding. This means that, for each patch, we would consider the different opinions of the pathologists to build a probabilistic vector for each pixel. For example, if a pixel was given grade 0 by one pathologist, grade 3 by two pathologists, grade 4 by three pathologists, and grade 5 by no one, the encoding would be $[1,2,3,0]/6=[0.166,0.333,0.5,0]$
\end{itemize}

Similar to DigestPath2019, we employed the same type of data augmentation and the following hyperparameters: categorical cross-entropy as loss function, nadam optimizer \cite{Dozat2016}, 250 iterations for one epoch (approximately the number of images), 400 epochs, an initial learning rate of $lr=0.001$, a rate decay of $lr_{decay}=0.99$, being the updated rate at each new epoch $lr_{new}= lr\times (lr_{decay})^{epoch}$, and no early-stopping.

\section{Results}

\subsection{DigestPath2019}

For the validation set, we obtained DICE of 82.49\%, whereas Timarisu network yielded a DICE of 80.84\%. To evaluate the importance of using a large patch, we tested the network with patches of 512$\times$512 pixels (doubling the GR=12, with 1.1M parameters) and 256$\times$256 pixels (setting the GR=24, with 4.3M parameters). The smaller patch resulted in a unstable, deficient training, with a DICE of 75.65\%. The 512$\times$512 patch yielded reasonable results, with a DICE 80.21\%. Therefore, it was beneficial to increase the patch size to 768$\times$768 even though that entailed to reduce the number of parameters to 294K. We also attempted to reduce the resolution of the images so that larger patches (1024$\times$1024 converted into 512$\times$512) could be used with bigger networks, but we observed that it affected the performance and quality of the results. Alternatively, we also tested reducing only the output image and later scaling it up, but that did not show any difference with respect to our propose setup. Eventually, the proposed balance between batch size and network depth was optimal considering the limit of 16 GB of GPU RAM.

Finally, we performed a 10-fold cross validation with the 1BN-DenseUnet (Fig.~\ref{fig03}) and we obtained the following average metrics: an accuracy of 99.93\% for the negative images, 94.76\% for the positive images, a total accuracy of 97.98\%, a F1-score of 79.17\%, and a DICE of 79.82\%, which was slightly lower than the challenge winner.

Qualitatively, we observed a very good segmentation (Fig.~\ref{fig05}):

\begin{itemize}
	\item Overall, the cancer areas were well detected in the positive images.
	\item Negative images were almost perfectly identified as non-cancer (99.93\% pixel accuracy).
	\item Some areas appeared blurred (mainly in positive images), which was an indication of complicated tissue morphology.
	\item The non-tissue areas were perfectly classified even though we built our batches without directly sampling from it.
	\item Our method tended to be more precise in non-tissue pixels that the gold standard wrongly indicated as cancer because of the way pathologist made the annotations (blue arrows and the whole B3 annotation in Fig.~\ref{fig04}).
	\item Some large non-cancer areas were classified as cancer with high certainty (green arrows in Fig.~\ref{fig04}), which made us wonder whether the annotations were correct in those cases. 
\end{itemize}

\subsection{Gleason2019}

\begin{figure}[t]
	\centering
	\includegraphics[width=1\linewidth]{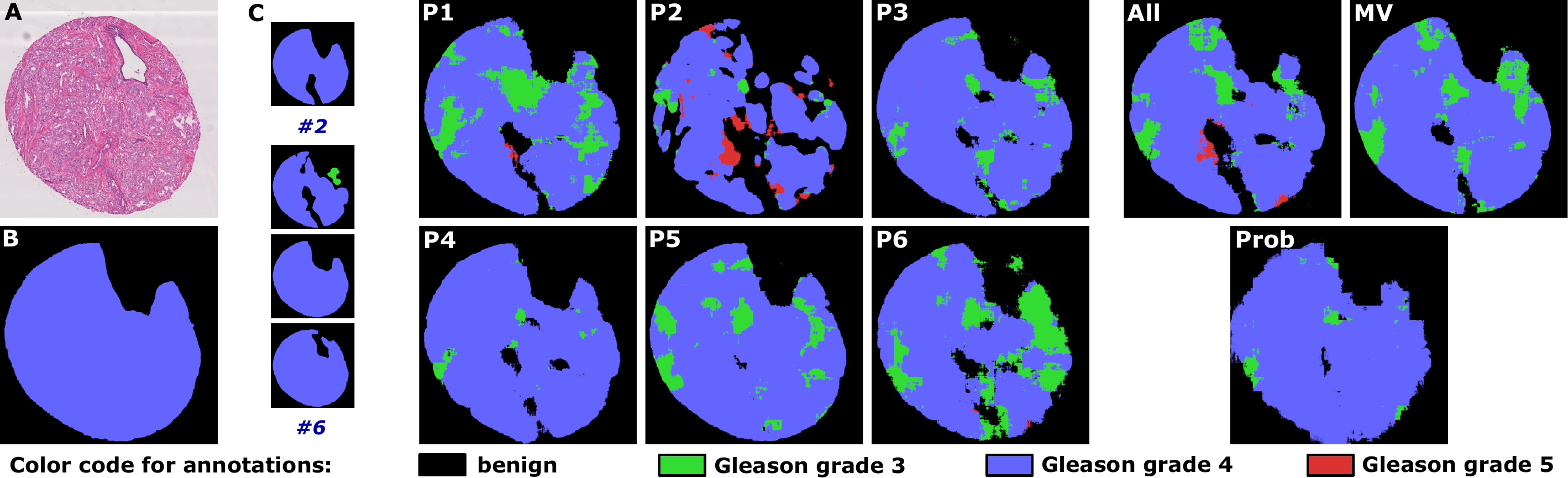}
	\caption{The output images for different setups in the Gleason challenge. \textbf{(A)} The prostate tissue image (input). \textbf{(B)} The majority vote annotation (label). \textbf{(C)} The annotated image from the pathologists; if an image was not annotated by one pathologist, their number is indicated instead (\#2 \& \#6). \textbf{(P1-P6)} The output for the setup where only the annotations of one pathologist were used for training. \textbf{(All)} The output using all annotations as training. \textbf{(MV)} The output using the majority-vote annotations as training. \textbf{(Prob)} The output using the probability approach.}
	\label{fig04}
\end{figure}

\begin{figure}[]
	\centering
	\includegraphics[width=1\linewidth]{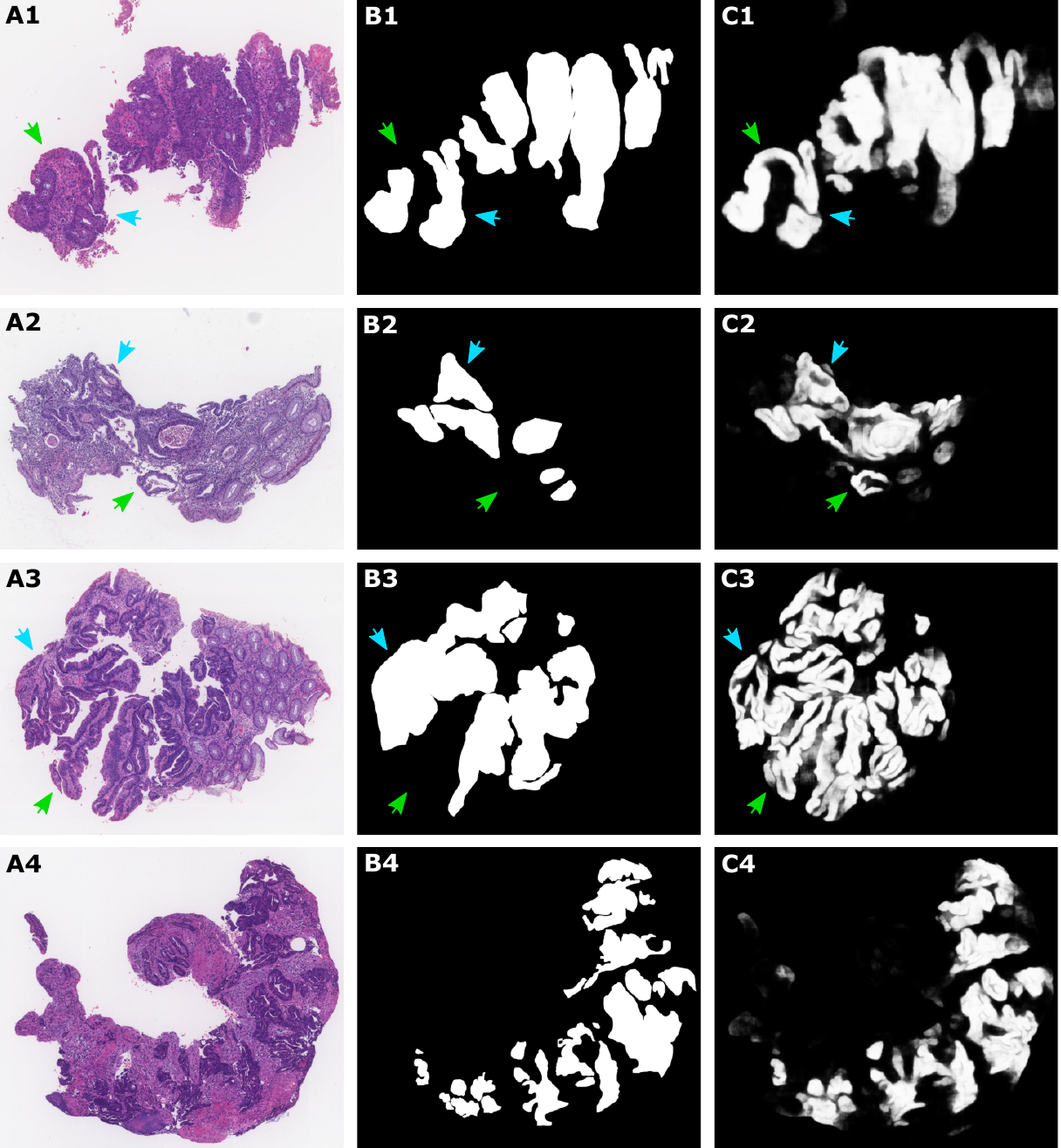}
	\caption{\textbf{(A)} Four representative examples of positive images with cancer tissue. \textbf{(B)} Pathologist's annotations (binary images). \textbf{(C)} Output of our network (probability images). Blue arrows indicate annotations poorly made (non-tissue areas within the cancer annotation). Green arrows indicate areas classified with high certainty as cancer that were annotated as non-cancer, which suggest possible mistakes in the pathologist's annotations.}
	\label{fig05}
\end{figure}

This dataset became a more complex problem due to the inconsistencies between pathologists. If the network was trained with the annotations of only one pathologist, the selection of the pathologist was key to a better performance (Table~\ref{table_Gleason}), being pathologists \#3 and \#5 the ones with greater accuracy. Since the evaluation was done with the majority-vote labels, this was not an indication of the quality of the pathologists' skills but an sign of which pathologists had similar annotation labels to the resulting majority-vote labels (Fig.~\ref{fig04}). 

We also tested an alternative setup where the network was trained and tested on the annotations of the same pathologist, and the results indicated that most pathologists scored higher in the majority-vote labels than in their own labels, with the exception pathologists \#2 and \#6 (Table~\ref{table_Gleason}). This could suggest the existence of inconsistent labels within the annotations of the same pathologist and the benefit of using the majority-vote as the best gold standard. More interestingly, the experiment where pathologist \#2 was used for training and testing was the only case where grade 5 had a non-zero DICE score (32.5\%), suggesting that he was precise and consistent in his annotations such that the network could differentiate all the Gleason grades, even with the existence of a high label unbalance.

As expected, training with the annotations of all pathologists was slightly detrimental, as random label sampling introduces contradictory annotations. In contrast, training with the majority-vote labels provided better performance, although this was not surprising (Table~\ref{table_Gleason}). Our proposed probability approach yielded the best categorical accuracy but not the best score  \footnote{Our score was considerably larger than the winner of the challenge, which made us wonder whether the metric could be wrongly defined. We attempted to contact the challenge organizers to clarify this, but without success.}. We observed that the probability approach was more sensitive to the `patch effect' in the output image, where the network was not capable of providing a smooth transition of grades between patches in the reconstructed output image (Fig.~\ref{fig04}). Finally, Tiramisu network yielded slightly worse results.

\setlength{\tabcolsep}{3.8pt}
\begin{table}
\begin{center}
	\caption{Categorical accuracy and score in the validation set for the different setups where the majority-vote was employed as test labels and using as training: only the annotations of one pathologist (P1--P6), all annotations (All), the majority-vote (MV), the probabilistic approach (Prob), or the Tiramisu network with probabilistic approach (Tir). The best metric in bold. For comparative purposes, each pathologist was also trained and tested on their \textbf{own} labels (bottom rows). }
	\label{table_Gleason}
	\begin{tabular}{lrrrrrrrrrr}
		\hline\noalign{\smallskip}
		& P1 & P2 & P3 & P4 & P5 & P6 & All & MV & Prob & Tir\\
		\noalign{\smallskip}
		\hline
		\noalign{\smallskip}
		No. images     &   242 &   136 &   238 &   240 &   244 &    36 &   244 &   244 &   244 &  244 \\[1pt]
		\hline 
		\\[-1em]
		Accuracy       & 78.58 & 74.84 & 85.21 & 84.09 & 85.22 & 76.42 & 78.99 & 85.56 & \textbf{85.66} & 85.14 \\[1pt]
		Score          & 1.192 & 1.104 & 1.335 & 1.299 & 1.356 & 1.132 & 1.224 & \textbf{1.359} & 1.343 & 1.311\\[1pt]
		\hline
		\\[-1em]
		Own acc.       & 73.70 & 83.05 & 82.40 & 82.36 & 84.87 & 78.75 & - & - & - & - \\[1pt]
		Own score      & 1.088 & 1.256 & 1.232 & 1.243 & 1.358 & 1.088 & - & - & - & - \\[1pt]
		\hline
	\end{tabular}
\end{center}
\end{table}
\setlength{\tabcolsep}{1.4pt}

\begin{figure}[]
	\centering
	\includegraphics[width=1\linewidth]{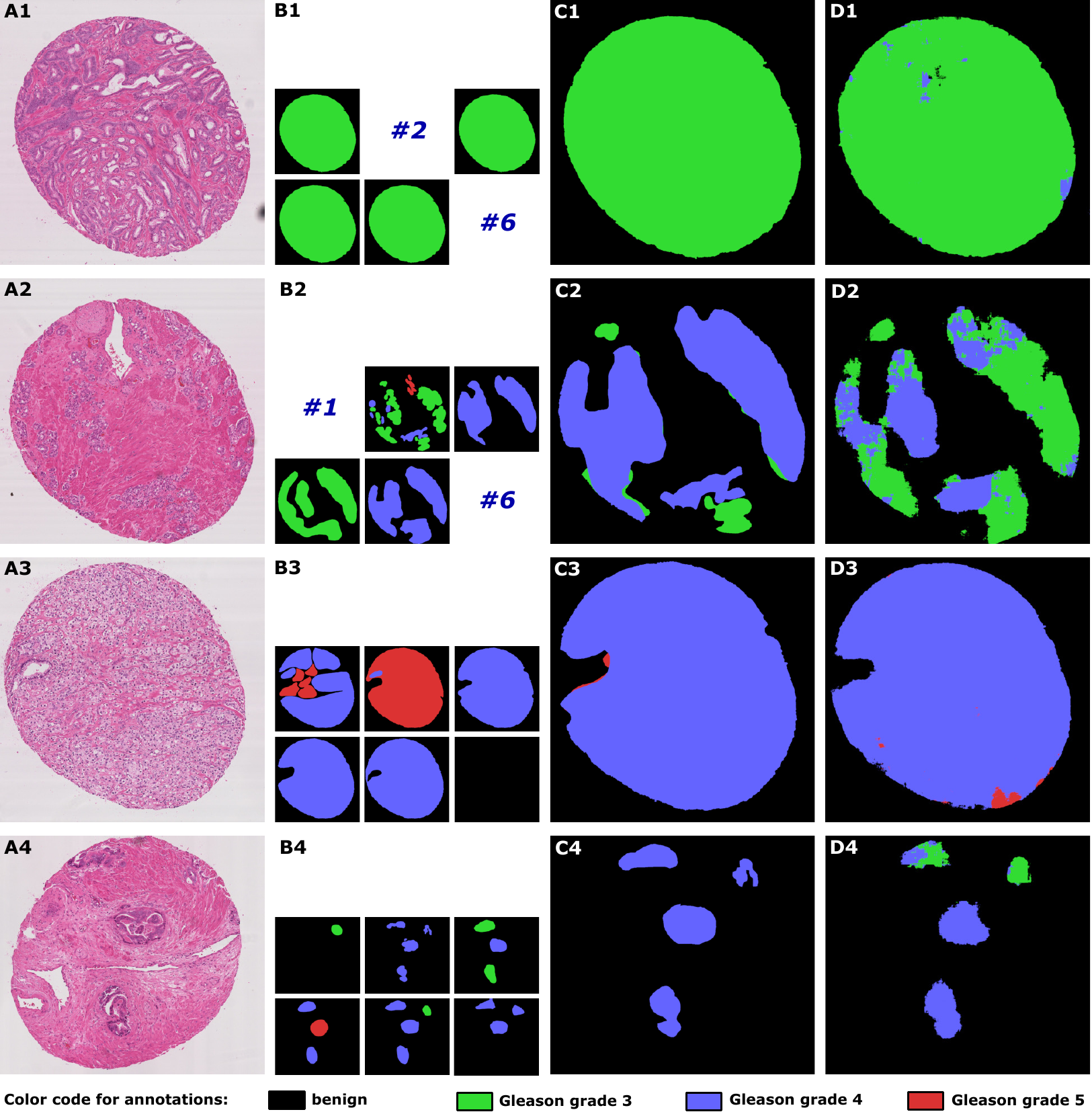}
	\caption{\textbf{(A)} Four representative examples of prostate tissue biopsies. \textbf{(B)} The annotated images from the pathologists; if an image was not annotated by one pathologist, their number is indicated instead (\#1,...). \textbf{(C)} The majority vote annotations; benign tissue (white pixels in Fig.~\ref{fig02}) was merged with non-tissue (black pixels). \textbf{(D)} Output of our network (categorical classification) training with the majority-vote labels. \textbf{(Bottom)} Color code.}
	\label{fig06}
\end{figure}

Qualitatively, our model could identify the areas with abnormal cells reasonably well, but the grading was sometimes inconsistent with the majority vote (Fig.~\ref{fig06}).

\section{Discussion}

In this work, we have presented a new DenseUnet that provides better performance than a Tiramisu network (with similar size) in two different histopathology problems, one being a two-class problem and another being a multi-class problem. These two datasets were presented as challenges in the MICCAI conference, which allows us to compare our performance with other teams. Indeed, our method, along with the training details that boost the network learning, provided a performance similar or better than the winners of the respective challenges, even though we did not add any fine-tune method specifically designed for one of the datasets since our goal was to present a generic network able to perform proficiently in different pathology problems.

In our experiments with the design of the network, we observed that including feature reduction layers in the beginning of the convolutional blocks (as originally designed in the DenseNets \cite{Huang17}) was beneficial, whereas Tiramisu network does not make use of them. BN layers were not crucial in our problem and they required many memory resources, so we reduced them to only the beginning of convolutional blocks instead of setting them before every convolutional layer. BN layers are largely used due to its benefits in accelerating the training while providing a small amount of regularization to the network \cite{Ioffe2015}. We actually did not observe any difference in convergence speed or performance if BN layers were completely removed from the network, but it had a small regularization effect and, thus, we kept it in the network. 

Our sampling method to build the batch, which consisted of selecting each patch from a different image, balancing the classes through that selection, and shuffling the order of the patches --avoiding the network to learn a specific distribution of classes in the batch--, had a large impact in the performance of the network. Indeed, if the patch sampling was reduced to a pair of positive-negative images (in DigestPath2019 dataset), the batch was not representative of the distribution of the whole dataset and the performance was highly affected. In this respect, batch renormalization (BRN) layers \cite{Ioffe2017} were effective in battling this problem, as the parameters used for normalization were computed over the different batches. Nevertheless, once we sampled from four different images, no differences were observed between a network with BN or BRN. Since BRN needed to store more parameters and thus using more memory, BN layers were preferred. Furthermore, we tested whether increasing the batch size would further improve the performance, although that entailed to reduce the network size (to fit in the GPU memory); our experiment suggested that more than four images were not particularly necessary as the reduction of the number of network parameters diminished the performance. 

We firmly believe that our network could learn the different cancer patterns and satisfactorily detected them (Fig.~\ref{fig05}). This was particularly manifest in the DigestPath2019 dataset, where in many cases our segmentation seemed to correct clear mistakes or imprecise delineations in the pathologists' annotations (blue arrows in Fig.~\ref{fig05}). Let us take the Fig.~\ref{fig05}-A3 as an example: the annotation was done in a rough, imprecise fashion, annotating many internal non-tissue areas as cancer; it could be argued that the presence of such non-tissue structure is an indication of cancer, but for a computer algorithm only the border of that area (which is the actual abnormal tissue structure) should be annotated. In other words, pathologists sometimes annotate images with the unconscious assumption that another human being would interpret their annotations or without having a basic understanding of how computer algorithms work. Given these inconsistencies, it is complicated to train a model that would be able to avoid being biased by the human mistakes in the annotations. In this respect, the visual analysis of our results made us believe our model was indeed robust against those inconsistencies.

Regarding the Gleason2019 dataset, our performance was rather suboptimal. By visual inspection (Fig.~\ref{fig06}), we could corroborate that our method was overall successful in detecting the areas with presence of cancer, but it did not perform so well in grading the tissue. However, it was unclear whether this was a network problem or a dataset/label problem. Indeed, the majority vote could be interpreted as the best gold standard if no other information is given. If the expertise of the pathologists were provided, it would probably be optimal to weight the labels based on their years of experience. This was already proven to be correlated with the accuracy of the annotations in the field of ophthalmology, where ophthalmologists with 25 years of experience were substantially better than new clinicians with 5-10 years of experiences \cite{Fauw18}. Nevertheless, we observed that the probabilistic approach was rather inferior qualitatively although not quantitatively. Therefore, we preferred the network that was trained with the majority-vote labels and one-hot encoding, although this case did not classify correctly the grade 5. Indeed, only pathologist \#2, who had very detailed annotations, performed very well and properly balancing the grades if the network was trained and tested on their own labels, which clearly suggested that it is important to make precise and detailed-oriented labels to yield unbiased results. It is worth noting that the majority-vote labels showed sometimes abnormal, illogical patterns, which would never be annotated in that way for any pathologist, but that was an unavoidable flaw of the methodology for evaluation.

\section{Conclusions}

We have proposed a Fully Convolutional DenseNet particularly designed for large histopathology images. We have shown that our network performs better than the well-known Tiramisu network \cite{Jegou17} for two different histopathology problems. These datasets were published in the MICCAI conference as two pathology challenges. Different from other natural images, histopathology images are considerably large, and the cancer patterns can reach large areas, therefore it is vital to solve these images with patches that cover such patterns. Given the currently limited resources in GPUs, it is important to adapt the default networks in the literature for this goal. Our proposed network, which was not fine-tuned to perform particularly better in one dataset, proved to be robust against different histopathology images, yielding similar or better results than challenges winners.

%\clearpage
\bibliographystyle{splncs04}
\bibliography{eccv2020_patho}

\begin{thebibliography}{10}
\providecommand{\url}[1]{\texttt{#1}}
\providecommand{\urlprefix}{URL }
\providecommand{\doi}[1]{https://doi.org/#1}

\bibitem{Clevert17}
Clevert, D.A., Unterthiner, T., Hochreiter, S.: Fast and accurate deep network
  learning by exponential linear units {(ELUs)}. In: International Conference
  on Learning Representations (ICLR) (2016)

\bibitem{Fauw18}
{De Fauw}, J., Ledsam, J.R., Romera-Paredes, B., {et al.}: Clinically
  applicable deep learning for diagnosis and referral in retinal disease.
  Nature Medicine  \textbf{24},  1342–--1350 (2018)

\bibitem{Dozat2016}
Dozat, T.: Incorportating {Nesterov} momentum into {Adam}. In: International
  Conference on Learning Representations (ICLR) Workshop. vol.~1, pp.
  2013--2016. (2016)

\bibitem{Drozdzal16}
Drozdzal, M., Vorontsov, E., Chartrand, G., Kadoury, S., Pal, C.: The
  importance of skip connections in biomedical image segmentation. In: Deep
  Learning and Data Labeling for Medical Applications. LNCS, vol. 10008.
  Springer, Cham (2016)

\bibitem{Glorot11}
Glorot, X., Bordes, A., Bengio, Y.: Deep sparse rectifier neural networks. In:
  Proceedings of the 14th International Conference on Artificial Intelligence
  and Statistics (AISTATS). vol.~15, pp. 315--323 (2011)

\bibitem{He2016}
He, K., Zhang, X., Ren, S., Sun, J.: Deep residual learning for image
  recognition. In: 2016 IEEE Conference on Computer Vision and Pattern
  Recognition (CVPR). pp. 770--778. Las Vegas, NV (2016)

\bibitem{Huang17}
Huang, G., Liu, Z., {van der Maaten}, L., Weinberger, K.Q.: Densely connected
  convolutional networks. In: 2017 IEEE Conference on Computer Vision and
  Pattern Recognition (CVPR) (2017)

\bibitem{Ioffe2017}
Ioffe, S.: Batch renormalization: towards reducing minibatch dependence in
  batch-normalized models. In: 31st Conference on Neural Information Processing
  Systems (NIPS). Long Beach, CA, USA. (2017)

\bibitem{Ioffe2015}
Ioffe, S., Szegedy, C.: Batch normalization: accelerating deep network training
  by reducing internal covariate shift. In: Proceedings of the 32nd
  International Conference on Machine Learning (ICML). vol.~37, pp. 448--456
  (2015)

\bibitem{Jegou17}
Jégou, S., Drozdzal, M., Vazquez, D., Romero, A., Bengio, Y.: The one hundred
  layers tiramisu: fully convolutional {DenseNets} for semantic segmentation.
  In: 2017 IEEE Conference on Computer Vision and Pattern Recognition Workshops
  (CVPRW). pp. 1175--1183. Honolulu, HI (2017)

\bibitem{Long15}
Long, J., Shelhamer, E., Darrell, T.: Fully convolutional networks for semantic
  segmentation. In: 2015 IEEE Conference on Computer Vision and Pattern
  Recognition (CVPR). pp. 3431--3440. IEEE, Boston, MA, (2015)

\bibitem{DigestPath19}
{MICCAI 2019}: {DigestPath Challenge}.
  \url{https://digestpath2019.grand-challenge.org}, [Online; accessed
  15-December-2019]

\bibitem{Gleason19}
{MICCAI 2019}: {Gleason Challenge}.
  \url{https://gleason2019.grand-challenge.org}, [Online; accessed
  15-December-2019]

\bibitem{scikitLearn}
Pedregosa, F., Varoquaux, G., Gramfort, A., Michel, V., Thirion, B., Grisel,
  O., Blondel, M., Prettenhofer, P., Weiss, R., Dubourg, V., Vanderplas, J.,
  Passos, A., Cournapeau, D., Brucher, M., Perrot, M., Duchesnay, E.:
  Scikit-learn: Machine learning in {P}ython. Journal of Machine Learning
  Research  \textbf{12},  2825--2830 (2011)

\bibitem{GleasonScore19}
{Prostate Cancer Foundation}: Gleason score.
  \url{https://www.pcf.org/about-prostate-cancer/diagnosis-staging-prostate-cancer/},
  [Online; accessed 10-February-2020]

\bibitem{Ronneberger15}
Ronneberger, O., Fischer, P., Brox, T.: {U-Net}: convolutional networks for
  biomedical image segmentation. In: Medical Image Computing and
  Computer-Assisted Intervention (MICCAI). vol.~9351, pp. 234--241 (2015)

\bibitem{ViguerasSPIE19}
Vigueras-Guill\'{e}n, J.P., Lemij, H.G., van Rooij, J., Vermeer, K.A., van
  Vliet, L.J.: {Automatic detection of the region of interest in corneal
  endothelium images using dense convolutional neural networks}. In: Medical
  Imaging 2019: Image Processing. vol. 10949, pp. 779--789. SPIE (2019)

\bibitem{Zeng2018}
Zeng, G., Zheng, G.: Multi-scale fully convolutional {DenseNets} for automated
  skin lesion segmentation in dermoscopy images. In: 15th International
  Conference Image Analysis and Recognition, ICIAR 2018. LNCS, vol. 10882, pp.
  513--521. Springer, Cham (2018)

\end{thebibliography}
\end{document}